# Applied Neural Cross-Correlation into the Curved Trajectory Detection Process for Braitenberg Vehicles

*Matin Macktoobian, Mohammad Jafari, Erfan Attarzadeh Gh.*
Computer Engineering Faculty
K. N. Toosi University of Technology
Tehran, Iran
matinking@hotmail.com

*Abstract* - Curved Trajectory Detection (CTD) process could be considered among high-level planned capabilities for cognitive agents, has which been acquired under aegis of embedded artificial spiking neuronal circuits. In this paper, hard-wired implementation of the cross-correlation, as the most common comparison-driven scheme for both natural and artificial bionic constructions named "Depth Detection Module (DDM)", has been taken into account. It is manifestation of efficient handling upon epileptic seizures due to application of both excitatory and inhibitory connections within the circuit structure. Presented traditional analytic approach of the cross-correlation computation with regard to our neural mapping technique and the acquired traced precision have been turned into account for coherent accomplishments of the aforementioned design in perspective of the desired accuracy upon high-level cognitive reactions. Furthermore, the proposed circuit could be fitted into the scalable neuronal network of the CTD, properly. Simulated denouements have been captured based on the computational model of PIONEER™ mobile robot to verify characteristics of the module, in detail.

## I. INTRODUCTION

It is a long time from seminal accomplishments within the neuroscience territory, as one of the most pioneering branches of the science, among human-beings. Primary steps by Cajal [1], were which led to glorious novel insights regarding to natural nervous system, not only was a breakthrough in the general formulization of the neural structures in the nature but established a robust path for future engineering efforts to imitate those natural mechanisms, artificially. Later, indispensable acquired points by Hugin and Huxley [2], corresponding to the precise dynamics of the neural tissues, prepared a deserving context for a systematic view over neural structures. A well-defined analysis upon dynamics of the neuronal swarms, especially in perspective of the human brain, has been presented by Haken [3], where one can follow concrete signal analysis in addition to permanent and transient outcomes of the firing neurons. The aforementioned obtained breakthroughs not only modified our attitude regarding the functionality of the natural nervous system but let us, as engineers, to be inspired deeply with the probable approaches for artificial realizations of such nature-driven bionic characteristics in order to culminate different bio-inspired engineering plans. Advent of some engineering fields like neuro-engineering and bio-robotics and their recent progressions might be beholden to the mentioned seminal discoveries.

As the other point of view, simulation and realization of various mental mechanisms of the human brain has been born concurrently, with the invention of the computer science and its interaction with the psychological knowledge. Planning [4], learning [5], autonomy [6], decision-making [7] are among the stuffs, are which under investigation of the researchers to be applied to artificial agents with due attention to biomimetic studies on the mental properties of our mysterious brain. In this case, the fully-smart robot might be considered as the creature, is which able to imitate cognitive and perceptive mental outcomes. One may notice the hard-wired perceptive implementations, thanks to their advantages in comparison with software-based strategies such as both trivial computational overhead and their noticeable operational speed. Where computer scientists almost barely take the algorithmic stuffs into account in order to reach into the high-level artificial cognitive interactions, praiseworthy approach might be the view, in which the intervention of the cybernetics has been applied into the case. The most well-known work has been done by pioneer cybernetician Braitenberg [8], where he introduced some primary proximity sensor-based creatures to imitate some emotional outcomes of our mind into autonomous robots. His success to simulate such high-level cognitive capabilities could be incredibly implied when one notice that pure programming-based schemes are not able to mimic these accomplishments, effectively, in spite of their noticeable computational complexity. Hence, closer attention of the research community notified the deserving denouements of the low-level implementation to gain high-level operation. For instance, Prahitar's struggles and his colleagues led to the optimization of the robot actions by consideration of the genetic algorithm [9]. Utilization of fuzzy techniques, might which lead to more flexibility within some field conditions, has been studied by [10] and [11]. An empirical navigation overview could be tracked by Yang's and Lee's researches and their teams in [12] and [13], respectively. Furthermore, Wang et al. [14] have focused on the mixed neuro-fuzzy idea for acquisition of better performances. Non-linear nature of the Braitenberg vehicles is the case which had not been taken into account until relatively recent Rano's researches [15]. Some other cybernetic ideas has been implemented by stressing on the classical control schemes, are which syncretized with biologic phenomena. In view of both the theoretical and empirical points, Navigation of mobile robots by neuromdulation and utilized weight updating schemes are studied by French [16].

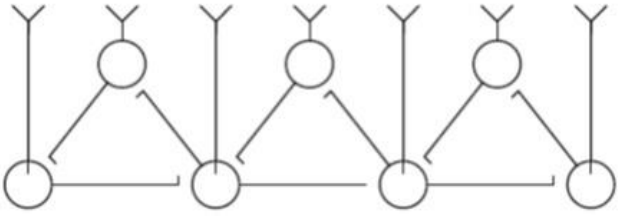

Fig. 1. Pure direction detector (PDD) module

One may pay close attention to the high-level outputs of internal Neuronal circuits to prepare the input for hyper high-level cognitive sub-systems. As a matter of fact, such more prolific information could be transferred to the next deductive layer of cognitive architecture. To this aim, Macktoobian [17] suggested a pure hardwired neuronal circuit based on the spiking model of the artificial neurons, to detect the direction of the wandering objects around the Braitenberg vehicles. Due to the exclusive capability of the mentioned design to detect straight bi-directional movements, the subsequent work [18] has been dedicated to take a novel circuit into account in order to detect motions of agents, in a more general form, i.e. curved movements, in which the tracked agent might either come closer, go farther or keep out of the straight trajectory. Considering modular structure of CTD, direction tracking is performed by PDD module, as shown in fig. 1, and depth detection has been realized by the other independent sub-circuit, has which been instantiated in a specific cascading formation. Thus, their optimization not only could be taken into account, one by one, but their raw outputs could be fed to the other detector systems. Fig. 2 might be captured to acquire an overall view upon CTD structure, in which the depth detection part needs to be determined, in detail.

In this paper, the sub-circuit corresponding to the depth detecting process will be analysed. As previously stated, the overall operation of the CTD must detect the relative motion of the target robot, is which, of course, within the active scope of the mounted proximity sensors on the host agent, such as the direction of its movement and the approaching state regarding to the reference robot.

The remainder of the paper is organized as follows. Section II presents a review upon the direction detection process, corresponding to the PDD module [18]. The neuronal circuit, so-called "Depth Detection Module (DDM)", is which designed in order to reveal the quality of the tracked curved trajectory, has been investigated in section III. On the computational aspects of the applied neural cross-correlation, one may refer to the section IV. Simulated accomplishments, to stress the credibility of the presented approach, could be pursued by contents of section V. At the end, section VI might be taken as a struggle into account in order to mention some generalized points about the neural manipulation in territory of cognitive robotics to conquer some facilitating attitudes for future researches, under aegis of the concluding remarks.

## II. DIRECTION DETECTION IN CTD PROCESS

As mentioned earlier, detective functionality of the motion direction has been considered to be performed separately regarding the depth evaluation. As described by Fig. 1, is which borrowed from the prior research [18]. A ternary circular set of neurons with inhibitory connections is responsible to react upon the direction of the motion corresponding to any wandering object around the embedded neurodetectors. Obviously, aggregation of such atomic units might lead to reach into the bigger PDD units, due to the scalable construction of the design. One also may notice the role of the inhibitory connections to handle the risk of the epileptic seizures, efficiently. Moreover, multi agent systems could be tracked with due attention to the fact that corresponding to each dynamic object, just one neuron within one atomic unit will be animated. Regarding to rational consideration of the relative big size of the objects in comparison with the scope of the detectors, every set of moving creatures could be tracked by this structure.

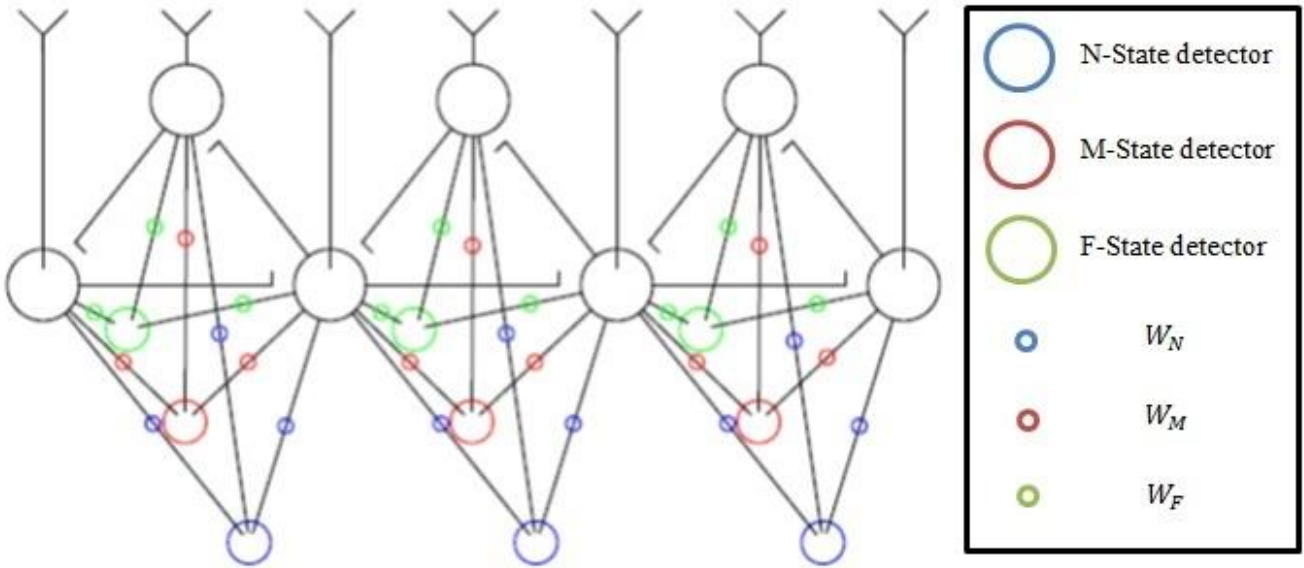

Fig. 2. Weight-tuned curved trajectory detector

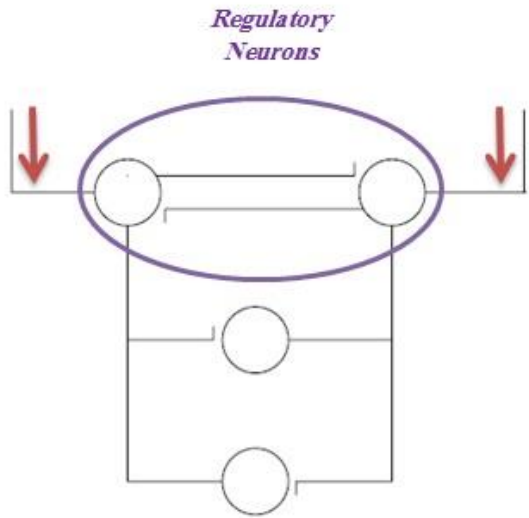

Fig. 3. Depth detection atomic unit

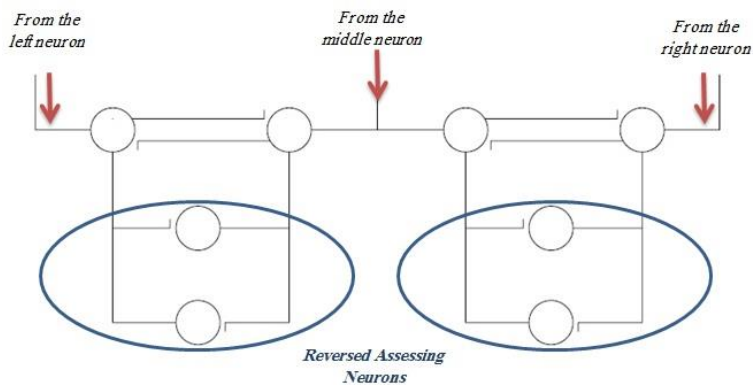

Fig. 4. Depth detection module (DDM)

The afterward step, would which need to be resolved, is the depth problem of our main partitioned problem. The previous rough criterion to distinguish among the 3 different possible holding states regarding to the CTD circuit, as are investigated in [18], elaborately, could be mapped into simple embedded weights. One can consider three judging neurons, connected to the neurodetectors, strengthened by those weights in order to be activated, exclusively, in view of the other 2 neurons. Even though, the described model is glorious in perspective of the simple analysis for highly-complicated generated spikes leading to the rough denouement regarding to the case, the resulting deficiencies upon the performance and even expecting modal robustness of the circuit are undeniable. As the main point, however one may assert that the epileptic concerns are managed within the PDD layer, the depth detection layer is barely constructed from the excitatory connections.

They could be incredibly challenging for large scale empirical implementations of the overall system. Leaving aside this, it is manifestation of the crucial desired functionality to tune the weight-ended neurons in sophisticated environments, with due attention to noticeable restricted DOFs, as 3 based on the number of the judging neurons. As a matter of fact, it will intrinsically lead to the obligation to utilize some other accessory neuronal levels to hit that mark.

Thus, one might look for the other engineering approach to not only eliminate above problematic points but solve the cognitive puzzle of the depth evaluation with taking some robust circuits into account. In very case, the new proposed technique, as a neuronal circuit, will be introduced in the next section. Notice that whatever CTD does, it must discern the approaching, straight motion and far-going conditions, in terms of the cognitive states, i.e. N-State, M-State and F-State, respectively, as the overall outcomes of the system.

III. DEPTH DETECTION MODULE

Just similar to the direction detection process, in which an atomic unit consisting a special structure of neurons, had been taken into account, it sounds inspiring to consider the same attitude upon the latter part of the CTD process, i.e. depth detection process. Therefore, an atomic unit for this aim might be depicted as fig. 3. The output signals of the direction detection level already have been introduced as the input signals for the depth detection process. Hence, one can mount two spike-driven signals from two neurons in the PDD circuit to feed into the depth detector circuit. The aforementioned signals, firstly, will compete upon activation of a reverse-pair of neurons. This stuff has been taken into account based on the fact that, within this neural race, each winning neuron will stimulate its own output, whereas the other one's output will be attenuated. With due attention to side-dependent structure of next level's sub-circuit, the mentioned competition and the acquired outcome will drive the circuit into different functionalities. In addition to such decision-making-based rule for the regulatory neurons, they are responsible for handling the probable epileptic seizures, such that the activated neuron in the neural competition will be made off by the expected activation of the other one within the next time slots. Hereupon, there are two single neurons, are which fed by output signals of the regulatory level, but in the reversed stimulating effect. Obviously, such closer the agent is to the neurodetecters, higher frequency upon the spike generation is expected. The reverse case is also credible, as well. As an instance, considering the left-to-right motion while the agent is approaching to the host, the right-sided neuron in the regulatory level will win the race and afterwards, the upper neuron in the assessing level will be activated. In the case of the bare straight trajectory in front of the neurodetectors, both of the assessing neurons will be deactivated. So, the M-State, according to the applied notation in [18], will be distinguished, properly.

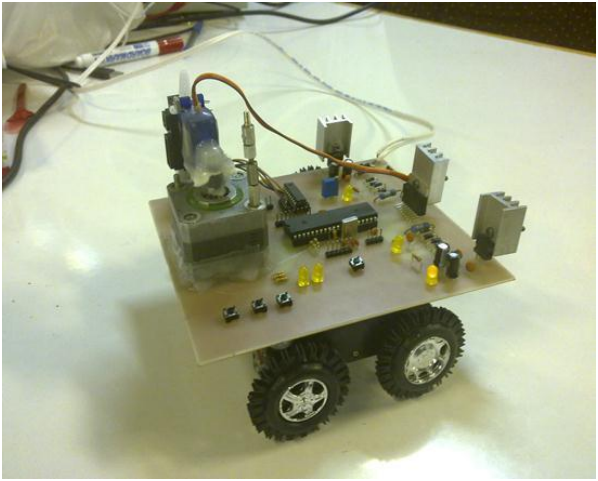

Fig. 5. PIONEER™ cognitive robot

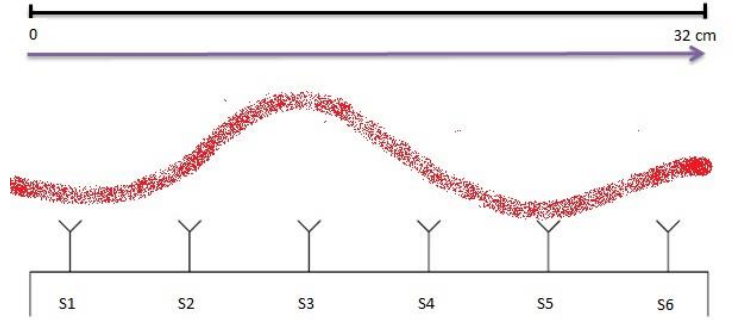

Fig. 6. Planned typical trajectory for wandering agent

All in all, one may admit that the described process is able to discern the relative depth of motion, regarding the applied stimulation into an adjacent pair of the neurodetectorrs. Of course, suffice it to say that the role of the assessing neurons is not unique in view of the either F-State or N-State. If the motion direction is changed, i.e. right-to-left approaching trajectory, the upper neuron will be responsible to act as the stressing factor to depict the F-State. The remainder of the design should be dedicated to the signals adaption between the two main modules, such as PDD and DDM. As PDD provides 3 outputs for the DDM, the adaptation condition could be sufficed with re-instantiating of the atomic depth module to mount between the other successive adjacent pair, corresponding to the PPD's neurons of the atomic unit, as shown in fig. 4.

## IV. SOME COMMENTS REGARDING THE NEURAL CROSS-CORRELATION

In analytic perspective, the interaction of the signals in the regulatory part of the DDM just associates the idea of the cross-correlation, in which high correlation among the aforementioned signals leads to deactivation of the assessing level, i.e. acquisition of the M-State; whereas the N-State and the F-State will be acquired by the trivial correlation of the named signals. Planning of the inhibitory connections in presence of the excitatory ones applies sign aspect to the spikes and transforms the traditional accumulative property of the cross-correlation to the algebric summation applied to the signed weighted spiking signals.

Generally, cross-correlation of two time-series could be acquired, as below:

$$Corr[X,Y]_w = \sum_{k=-\infty}^{\infty} X_k Y_{k+w} \quad (1)$$

Where, raw signals are shown by $X$ & $Y$. Furthermore, $w$ stands for the multiplying window's size.

As mentioned earlier, signed version of the above formula sounds more proper to acquisition of the both positive and negative interactions between the spiking signals, are which generated by the regulatory neurons and fed into the assessing neurons. A typical suggestion for realization of the neural cross-correlation based on the excitatory and inhibitory factors might be considered as below:

$$NCorr[X,Y]_w = \sum_{k=-\infty}^{\infty} X_k^{E \oplus I} Y_{k+w}^{E \oplus I} \quad (2)$$

Where $E \oplus I$ has been taken into account as a realization of exclusive selection between either excitatory or inhibitory connections. It is obvious that inhibitory option will change the signal sign to the negative form.

## V. SIMULATED OUTCOMES

Numerical simulations and practical tests are actually expected to reveal operational aspects of the new CTD and clarify the mentioned stipulations about the functionalities, more clearly. Thus, this section has been planned to capture some incredible improvements in view of the smooth potential variation of the CTD circuit in presence of the DMM, in comparison with the similar test, where the circuit is equipped with the tuned weights. Computational model of the PIONEER™ mobile robot, shown as fig. 5, has been used as a test bed to assess the credibility of the designed strategies under aegis of the simulated results.

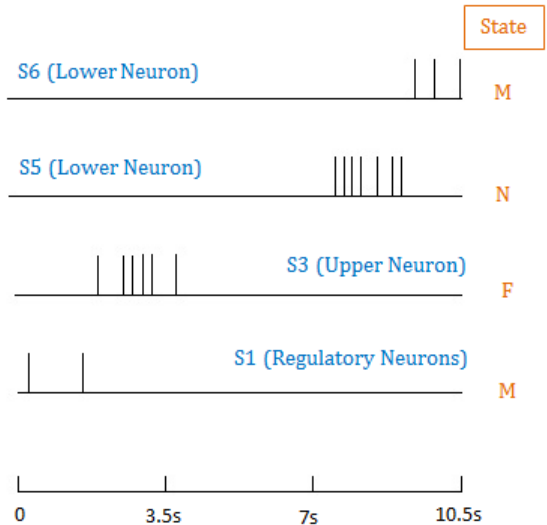

Fig. 7. Monitored generated spikes by DDM

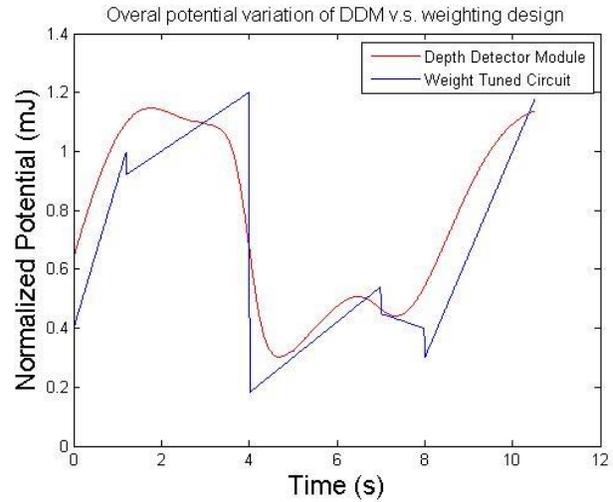

Fig. 8. Potential variation of DDM versus. weighting design

The mentioned wheel mobile robot has been equipped with 4 IR transducers as artificial neurodetectors; But for acquisition of more coherent data sets, computational model of the robot has been evolved and the simulations are taken into account under study of a 6-sensor version of PIONEER™. Mounted sensors had been planned to generate spike trains in presence of the dynamic objects, wandering in their active scope of sensitivity. A typical range-scaled trajectory has been considered just like fig. 6. Typical spike trains referring to neurons of the some DMMs have been demonstrated, as shown in fig. 7. One might pay attention to the potential level, has which not been decreased, noticeably with utilization of the DMM in comparison with the using weight neurons, is which depicted in fig. 8. It is due to the overall increment of the number of the neurons, as the DMM has been added to the circuit. But gratifying denouement in view of the potential variation is undeniable with due attention to the acquired smooth dynamics of the potential level.

## VI. CONCLUSION AND FUTURE WORKS

Current research is an attempt to improve the functionality of the Curved Trajectory Detection (CTD) process in both perspectives of structural and functional outcomes. Active manipulation of the neural signals, are which generated by Pure Direction Detector (PDD) part, by Depth Detector Module (DDM) has been considered as a robust technique to distinguish among different mutual exclusive cognitive states of the system. In the other words, cross-correlation, as a well-known procedure for comparing stuffs among the signals, has been captured for imitation of a similar scheme to be realized in the context of the neuronal circuits. Embedded DDM not only strengthens the DOFs of the circuit in order to acquire more flexibility for future progressions, but also leads to gain some improvements in comparison with the previous CTD circuit, is which based on the bare weight neurons. Mixed utilization of the excitatory and inhibitory connections is a witness for better handling of the epilepsy within the crucial time slots of the system operation. Moreover, alleviation of the potential variation and decreasing the intense rates of the potential might be noted as gratifying points about the novel CTD.

One may be prone to take some enlightening aspects into account just like key points for the future research and inspiration upon either acquiring deeper studies about DDM-driven CTD or new generations of the CTD and the other neurocognitive circuits. Estimation of the delays and concrete timing analysis of the circuit sounds to be noteworthy. There might be some ideas for transforming the proposed DDM to a unary circuit in order to avoid re-instantiating. Analytic investigation on the neural cross-correlation and try to establish better definitions with higher performance, especially in view of the computational aspects, might be followed.